\documentclass{article}

\usepackage{microtype}
\usepackage{graphicx}
\usepackage{booktabs} 
\usepackage{subcaption}

\usepackage{hyperref}

\usepackage[accepted]{icml2024}

\usepackage{amsmath}
\usepackage{amssymb}
\usepackage{mathtools}
\usepackage{amsthm}

\usepackage[capitalize,noabbrev]{cleveref}

\theoremstyle{plain}

\theoremstyle{definition}

\theoremstyle{remark}

\usepackage[textsize=tiny]{todonotes}

\icmltitlerunning{Information-Theoretic Progress Measures reveal Grokking is an Emergent Phase Transition}

\begin{document}

\twocolumn[
\icmltitle{Information-Theoretic Progress Measures reveal Grokking is an Emergent Phase Transition}



\icmlsetsymbol{equal}{*}

\begin{icmlauthorlist}
\icmlauthor{Kenzo Clauw}{ghent}
\icmlauthor{Sebastiano Stramaglia}{bari}
\icmlauthor{Daniele Marinazzo}{ghent}
\end{icmlauthorlist}

\icmlaffiliation{ghent}{Department of Data Analysis, Ghent University, Belgium}
\icmlaffiliation{bari}{Physics Department, University of Bari, Italy}

\icmlcorrespondingauthor{Kenzo Clauw}{kenzoclauw@gmail.com}
\icmlcorrespondingauthor{Sebastiano Stramaglia}{sebastiano.stramaglia@uniba.it}
\icmlcorrespondingauthor{Daniele Marinazzo}{daniele.marinazzo@ugent.be}

\icmlkeywords{Machine Learning, ICML}

\vskip 0.3in
]

\begin{abstract}
This paper studies emergent phenomena in neural networks by focusing on grokking where models suddenly generalize after delayed memorization. To understand this phase transition, we utilize higher-order mutual information to analyze the collective behavior (synergy) and shared properties (redundancy) between neurons during training. We identify distinct phases before grokking allowing us to anticipate when it occurs. We attribute grokking to an emergent phase transition caused by the synergistic interactions between neurons as a whole. We show that weight decay and weight initialization can enhance the emergent phase. 
\end{abstract}
\section{Introduction}





Grokking is a phenomenon where neural networks during training suddenly generalize after prolonged memorization with limited progress in the loss. Understanding this phase transition is vital for AI safety and alignment. While \cite{wei_emergent_2022} claim these models exhibit emergent properties, others attribute it to the choice of evaluation metric \cite{schaeffer_are_nodate}. This lack of consensus poses a key question: Are neural networks internally learning emergent behavior? 

To understand and predict emergent phase transitions, one approach is to identify hidden progress measures – metrics causally related to the loss \cite{barak_hidden_2023} - to predict and understand grokking. Recent work found progress measures based on mechanistic interpretability by reverse engineering models into interpretable components (i.e. circuits of subnetworks) \cite{nanda_progress_2023}. However, this approach has several limitations. First, the reliance on human effort to identify circuits prevents scaling to larger models and it is susceptible to subjective bias. Second, these measures are task-specific. Lastly, it ignores the emergent interactions between components.


This work proposes information theory as a task-independent tool to identify emergent sub-networks in neural networks for mechanistic interpretability. Pairwise mutual information has been shown to identify important features \cite{liu_understanding_2018}. However, it can only measure the dependency between two variables. Multivariate mutual information provides a fine-grained analysis of the statistical interactions between multiple variables by decomposing the dependencies into synergy and redundancy. Synergy refers to the cooperative behavior between variables as a whole, where their combined statistical interactions exceed the sum of their contributions in isolation. Redundancy is the shared information between variables. Motivated by recent work identifying synergistic sub-networks for feature importance \cite{clauw_higher-order_2022}, we study higher-order interactions between neurons to understand grokking.

In this paper, we hypothesize that grokking is a phase transition caused by the emergence of a generalizing sub-network due to the collective interactions between neurons as a whole, which cannot be quantified using pairwise metrics. To illustrate this, we study the simplest setting where grokking is observed - fully connected neural networks on modular arithmetic - as a case study for emergence. To understand grokking, we utilize the O-Information - a multivariate information theory measure that scales to multiple variables - to quantify the synergy and redundancy in a network \cite{rosas_quantifying_2019}. We study two strategies that impact grokking: weight decay and high initialization of the weights \cite{liu_omnigrok_2022}. Our contributions are as follows :
\begin{itemize}
     \item Using the synergy and redundancy as progress measures, we identify three key phases during training: \textbf{Feature Learning} of low-level patterns, \textbf{Emergence} of a generalizing sub-network, and \textbf{Decoupling} for compression.
    \item We observe that a low weight decay value results in an additional \textbf{divergent} and \textbf{delayed} \textbf{emergence} phase. We find that weight initialization and increasing weight decay directly leads to an emergent phase and reduces the delayed generalization gap
    \item We present preliminary findings indicating that the sub-networks at the emergent phase may be causally related to delayed generalization.
    \item We show that early peaks of synergy can predict if grokking occurs
\end{itemize}

\section{Related Work}
\textbf{Cause of Grokking} Several studies attribute the cause of grokking to properties related to the difficulty of representation learning the generalizing sub-network, including delayed feature learning \cite{kumar_grokking_2023}, sparsity \cite{merrill_tale_2023}, and compression 
 \cite{liu_grokking_nodate}. While these properties are observed in emergent complex systems, these interpretations do not consider the relationships between components as a whole. 

\textbf{Progress Measures} Several progress measures have been proposed based on L2 norm \cite{liu_omnigrok_2022} or Fourier gap \cite{barak_hidden_2023}. However, these are mostly heuristics. One study uses pairwise mutual information for measuring progress \cite{tan_understanding_2023}. None of these metrics quantify higher-order interactions between neurons.

\section{Methodology}
\textbf{Setup} We study the modular addition operation $\mathbb{Z}_p$ by generating equations of the form  $(a + b) \% p = c$ where $p$ = 97. To reduce confounding factors, we study a simple 2-layer fully connected network with ReLU activation functions. The input $(a, b) \in \mathbb{Z}_p \times \mathbb{Z}_p$ is a vector $x \in \mathbb{R}^{2p}$ represented by the concatenation of the one-hot representations of $a$ and $b$. Our model processes the input as $f(\theta; x) = W_2 Z^{1}$ where $Z^{1} = \text{ReLU} (W_1 x + b_1) \in  \mathbb{R}^{s \times n}$ is a matrix of feature vectors $z_{1}^{1}, .., z_{n}^{1}$ in the first hidden layer where $n$ = 250 is the number of neurons in the layer, and each feature vector $z_{n}^{1}$ has a dimension equal to the number of samples $s$ in the input data. For each experiment, we train 5 models with different random seeds using full-batch AdamW optimization with a 0.03 learning rate and 40 \% of the data \cite{loshchilov_fixing_2018}. 

\textbf{O-Information} Given a collection of $n$ random variables  $\mathbf{Z} = \{X_{1}, .., X_{n}\}$, the O-Information \cite{rosas_quantifying_2019} is defined as :
\begin{equation}
\Omega_{n}(\mathbf{Z}) = (n - 2)H(\mathbf{Z}) + \sum_{j=1}^{n} [H(Z_{j}) - H(\mathbf{Z} \backslash Z_{j})]
\label{eq:1}
\end{equation}
where $H$ is the entropy, and $\mathbf{Z} \backslash Z_{j}$ is the complement of $Z_{j}$ with respect to $\mathbf{Z}$. If $\Omega_{n}(\mathbf{Z}) > 0$ the system is redundancy-dominated, while if $\Omega_{n}(\mathbf{Z}) < 0$  it is synergy-dominated. We estimate the entropy terms in \ref{eq:1} using Gaussian Copula transformation \cite{ince_statistical_2017} allowing an efficient closed-form solution (see papers for details). 

\textbf{Quantify synergy and redundancy} We first reduce the dimensions of the features $Z^{1} \in \mathbb{R}^{s \times 250}$ by grouping similar feature vectors into 10 bins using agglomerative clustering based on the standard configuration in sklearn resulting in a feature matrix $\tilde{Z}^{1} \in \mathbb{R}^{s \times 10}$ where each bin in $\tilde{Z}^{1}$ consist of a set of similar feature vectors $\tilde{z}_{1}^{k}$. To quantify the synergy and redundancy in a network, we perform an exhaustive search of each combination of multiplets (i.e. subsets of 2 to $k$ bins where the largest combination $k$ = 10) using the O-Information $\Omega_{k}(\tilde{z}_{1}^{1}, .., \tilde{z}_{1}^{k})$ to find the optimal synergy (lowest) and redundancy (highest) value. In our plots, we normalize both synergy and and redundancy between 0 and 1 while inverting the synergy for comparison.

\section{Evaluating grokking}
\begin{figure}[ht]
\begin{center}
\centerline{\includegraphics[scale=0.25]{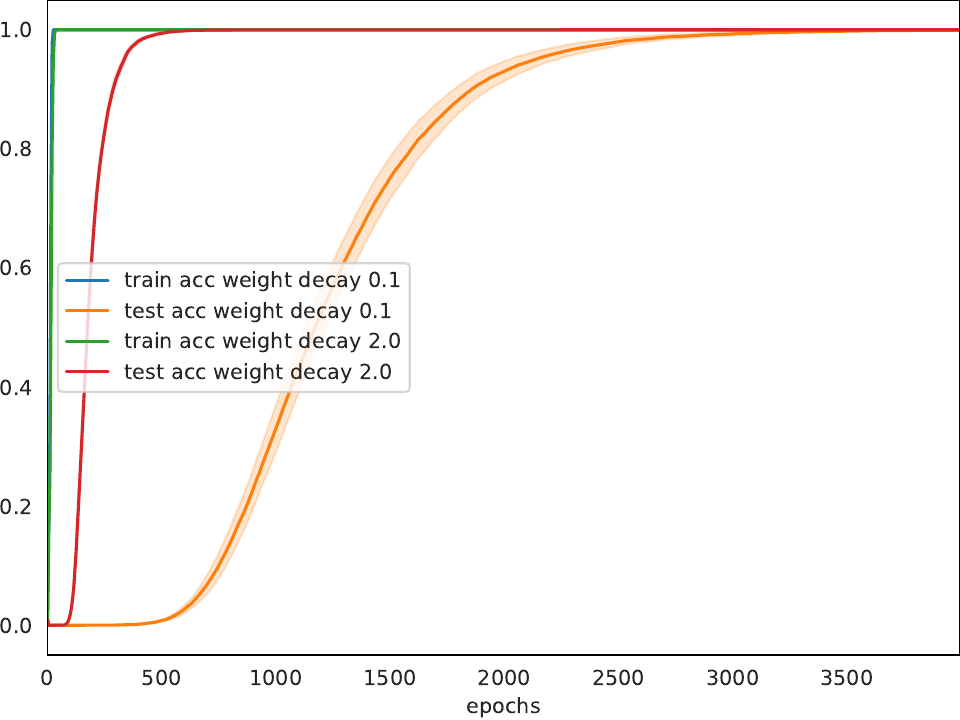}
            \includegraphics[scale=0.25]{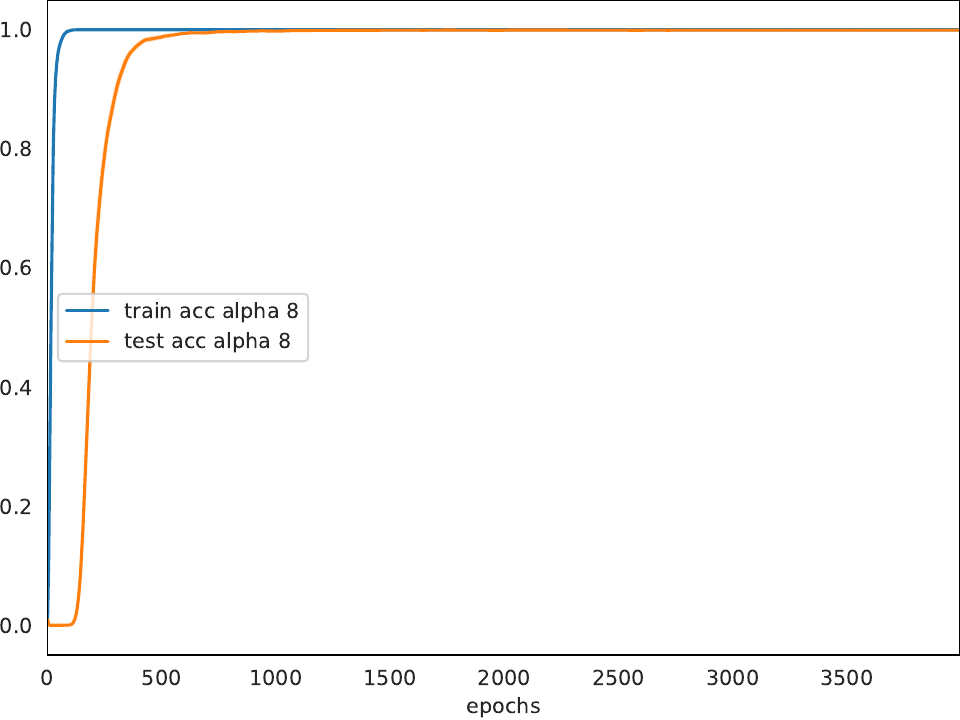}}
\caption{Left: accuracy for weight decay 0.1 and 2.0, Right: accuracy for alpha initialization 8.}
\label{fig:1}
\end{center}
\vskip -0.4in
\end{figure}

In this experiment, we evaluate the impact of design choices to simulate grokking by training models according to two strategies:

\textbf{Role of weight decay} Figure \ref{fig:1} illustrates models trained with low (0.01) and high (2.0) weight decay. We observe that a low weight decay results in delayed generalization. However, increasing weight decay reduces the generalization gap and increases its sharpness. 

\textbf{Role of initialization} To increase the validity of our findings, we investigate grokking beyond weight decay. \cite{liu_omnigrok_2022} observed that scaling the initialization weights of the model with a factor $\alpha > 0$ 
while constraining the L2 weight norm to be constant during training can induce grokking. Here, we train a model with a high initialization factor ($\alpha$ = 8) with zero weight decay. Similar to \cite{lyu_dichotomy_2023}, we set the initialization weights in the last layer to zero to avoid learning instabilities when using a large $\alpha$. Figure \ref{fig:1} (right) verifies that this strategy reduces the generalization gap while increasing its sharpness.

\section{Grokking is an emergent phase transition}
To understand grokking, we plot the normalized synergy and redundancy with the loss, as this reflects the training dynamics,  to measure the model's progress during training. Each measure is on a log scale for visualization purposes. We include Pareto plots to illustrate the trade-off between synergy and redundancy as this provides a more fine-grained analysis of the progress measures.

\subsection{Low weight decay delays the emergence phase}
We first study the most common setting of grokking for a model trained with low weight decay of 0.1. From figure \ref{fig:1} (left) we observe that training can be classified into 5 distinct phases:

\textbf{Feature Learning} We initially observed low synergy and high redundancy. The lack of coupling between features suggests the network is independently learning features, while the high redundancy indicates it is extracting basic patterns with similar properties.

\begin{figure}[ht]
\vskip 0.1in
\begin{center}
\centerline{\includegraphics[scale=0.25]{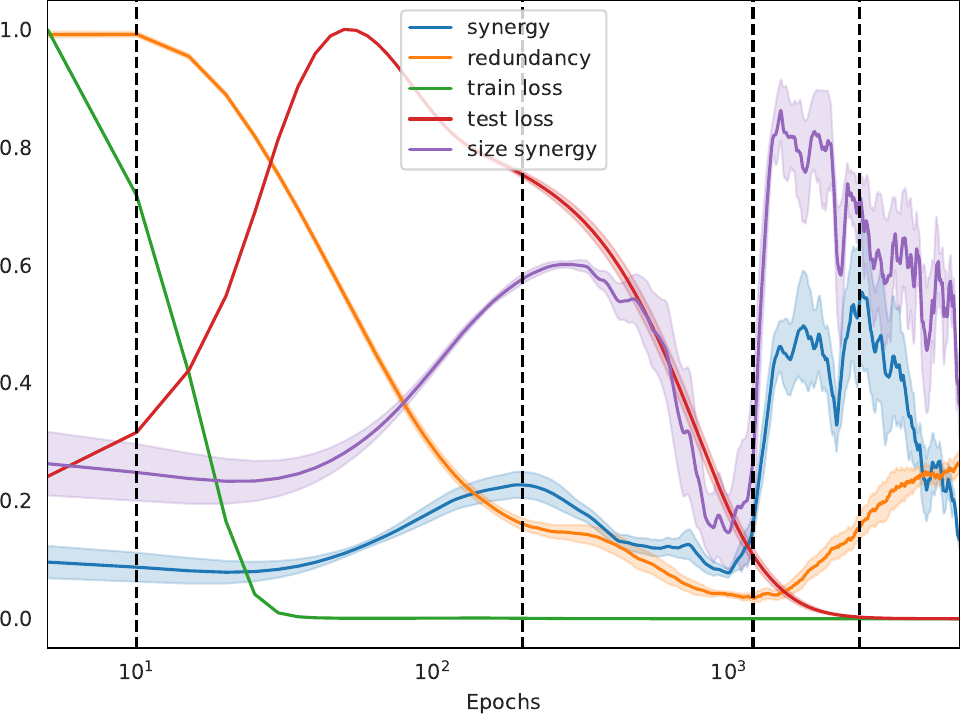}
            \includegraphics[scale=0.25]{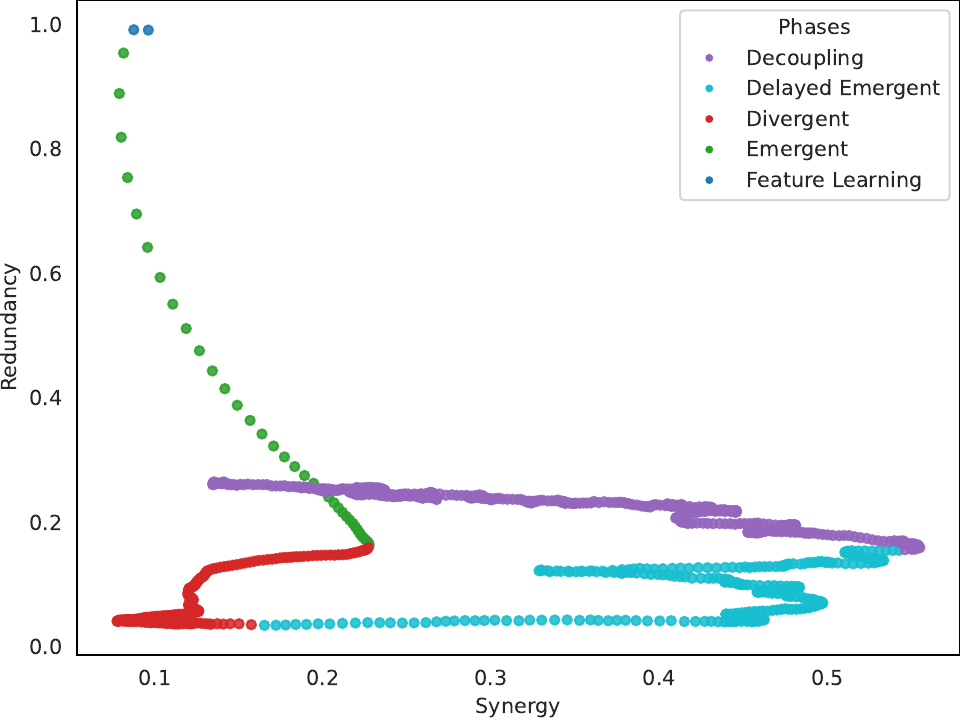}}
\caption{Baseline model with weight decay 0.1}
\label{low_weight_decay}
\end{center}
\vskip -0.2in
\end{figure}

\textbf{Emergence} During this phase, the Pareto front in figure \ref{low_weight_decay} (right) reveals that the model rapidly trades off redundancy for synergy. At the same time, the size of the synergistic sub-network increases, and a peak is observed in the test loss. This suggests a critical phase transition from memorization where the model attempts to combine features to emerge a generalizing synergistic sub-network that is growing as it explores.

\textbf{Divergence}
For low weight decay, the emergent phase does not directly lead to a generalizing solution. During a divergent phase, both the synergy and redundancy of the model drop, while the synergistic sub-network decreases in size with a small peak in the test loss. This indicates the model is overfitting due to a lack of sufficient features to combine. We hypothesize this is due to a low weight decay resulting in a difficult loss landscape making it difficult to learn generalizing. 

\textbf{Delayed Emergence} During this phase, we observe a rapid increase in synergy, redundancy, and the size of the synergistic sub-network. This suggests the model escaped the sub-optimal solution and is now able to combine features to form a generalizing solution.
\\\\ \textbf{Decoupling} In this phase, synergy decreases, redundancy increases, and test accuracy increases. This suggests the model has found a generalizing sub-network but not all features are relevant so it removes coupled interactions for compression. 
\subsection{Increasing weight decay enhances emergence}
\begin{figure}[ht]
\vskip 0.1in
\begin{center}
\centerline{\includegraphics[scale=0.25]{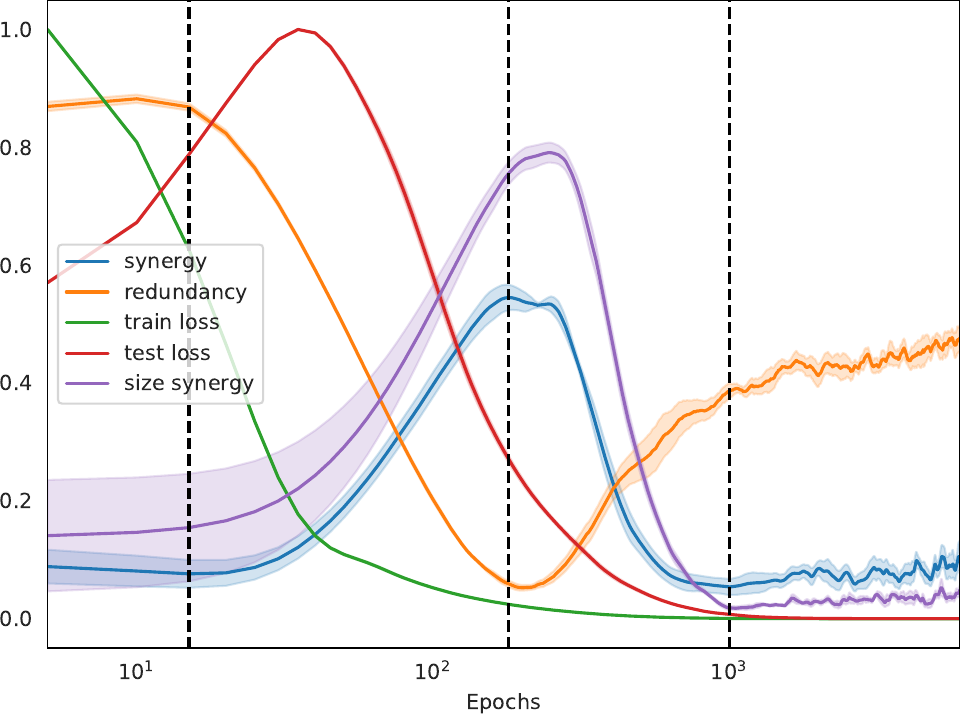}
            \includegraphics[scale=0.25]{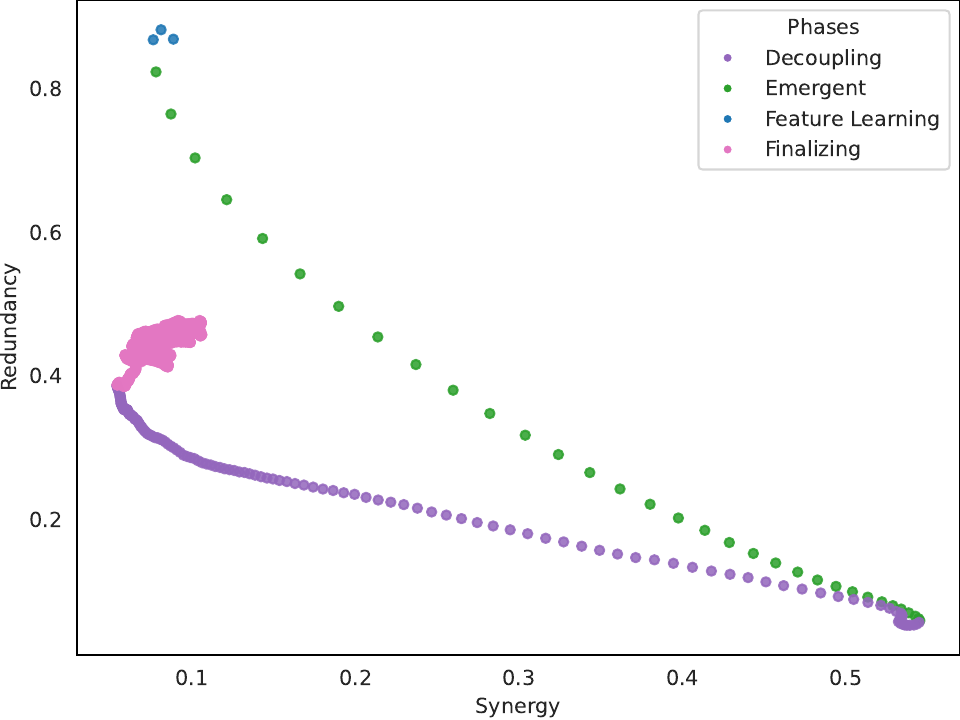}}

\caption{Baseline model with weight decay 2.0}
\label{high_weight_decay}
\end{center}
\vskip -0.2in
\end{figure}

From Figure \ref{high_weight_decay} (left), we observe that increasing weight decay directly results in an emergent phase with a larger peak in synergy and size of the synergistic sub-network. Moreover, the decoupling phase occurs immediately after this phase, with no divergent and delayed emergence phase. 

This suggests that weight decay acts as regularization either by reducing the capacity of the network, which promotes learning shared features that are more robust by reducing the number of active features and thereby decreasing overfitting, or by smoothing the loss landscape, making it easier for optimization to find regions where more features can be combined. 

We additionally observe a finalizing phase with minimal synergy and redundancy changes after test loss convergence. We attribute this to compression of the representation which is interesting for transfer learning but is not explored in this work. 

\subsection{The role of weight initialization}
\begin{figure}[ht]
\vskip 0.1in
\begin{center}
\centerline{\includegraphics[scale=0.25]{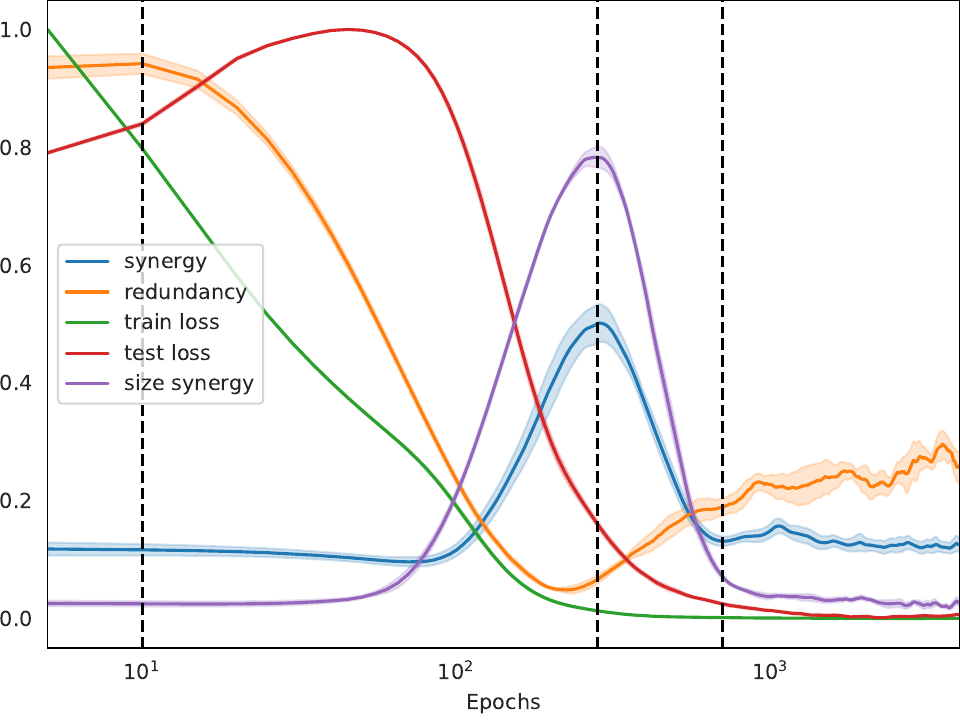}
            \includegraphics[scale=0.25]{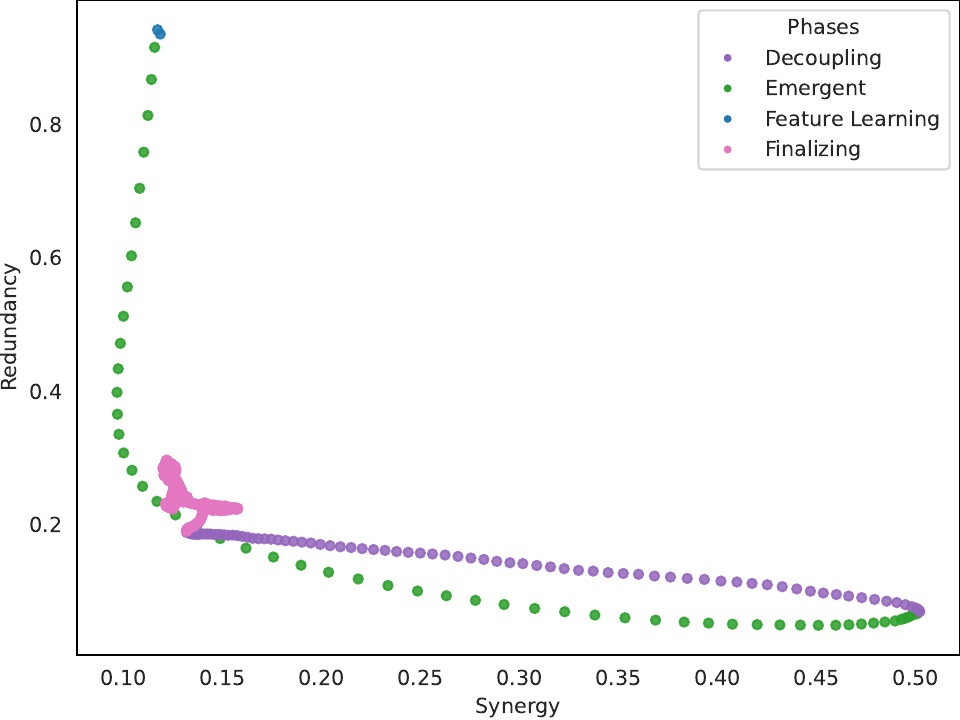}}

\caption{Baseline model with alpha = 8 and weight decay = 0}
\label{alpha}
\end{center}
\vskip -0.2in
\end{figure}

To isolate the contribution of weight decay regularization, we evaluate a model trained without weight decay using high weight initialization and a constrained norm. Figure \ref{alpha} (left) illustrates that a model trained with high weight initialization without weight decay directly results in an emergent phase. However, from Figure \ref{alpha} (right) we initially observe a rapid drop in redundancy followed by a delayed rapid increase in synergy. We hypothesize that this observation is due to the high convexity of the loss landscape making it easier for optimization to find a solution \cite{vysogorets_deconstructing_2024, fort_goldilocks_2018}. 
\section{Early synergy peaks predict grokking}
In this experiment, we plot the synergy for a variety of models trained with weight decay and weight initialization. We empirically provide evidence that the synergy can predict grokking if there is a small peak early in training. For comparison, we additionally train the models with the same parameters from section 4 but with higher values for both weight decay and high weight initialization. 

\textbf{Weight Decay} We additionally train models with weight decay values of 10 and 50, which do not achieve grokking (see figure 9 in the appendix).  From figure \ref{peak} (left) we observe that both low (0.1) and higher (2) weight decay have a peak prior to grokking occurring. However, if weight decay is too high (i.e. 10 and 50) then the model is not able to grok indicated by lower synergy. 

\textbf{Initialization} We additionally train models with weight decay values of 10 and 50, which do not achieve grokking (see figure 10 in the appendix). From figure \ref{peak} (right) we observe that high weight initialization (8) results in a synergistic peak. However, if alpha is too low (1) or too high (5) then the synergy rapidly drops and the model never recovers. 

These results indicate that early synergy peaks might be indicative of grokking. However, further experimentation with a variety of models and statistical tests is necessary to verify this hypothesis.

\begin{figure}[ht]
\vskip 0.1in
\begin{center}
\centerline{\includegraphics[scale=0.25]{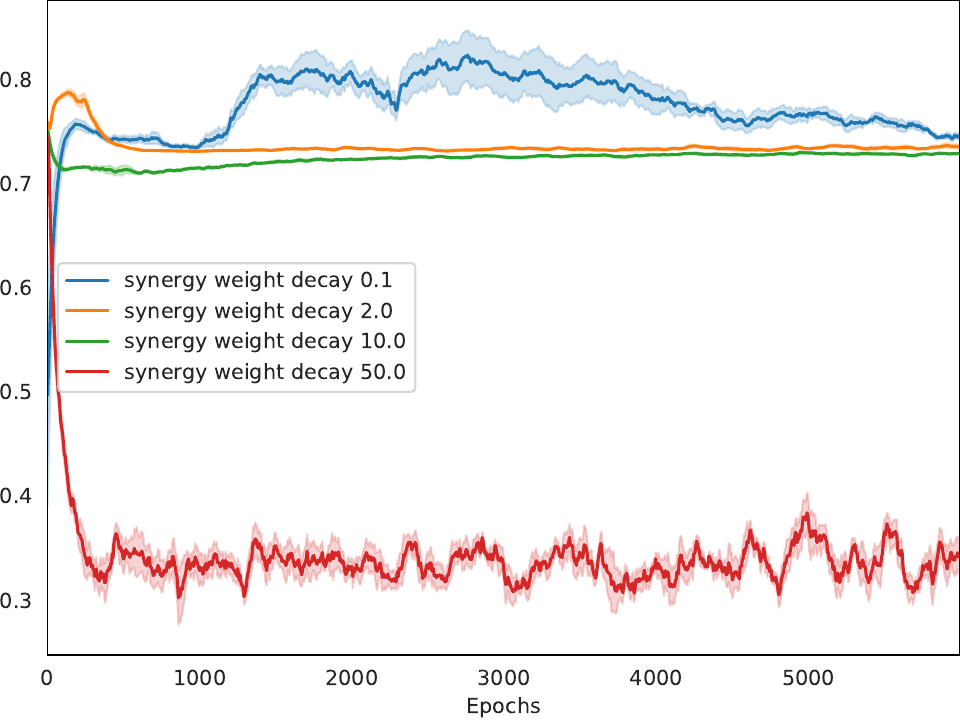}
            \includegraphics[scale=0.25]{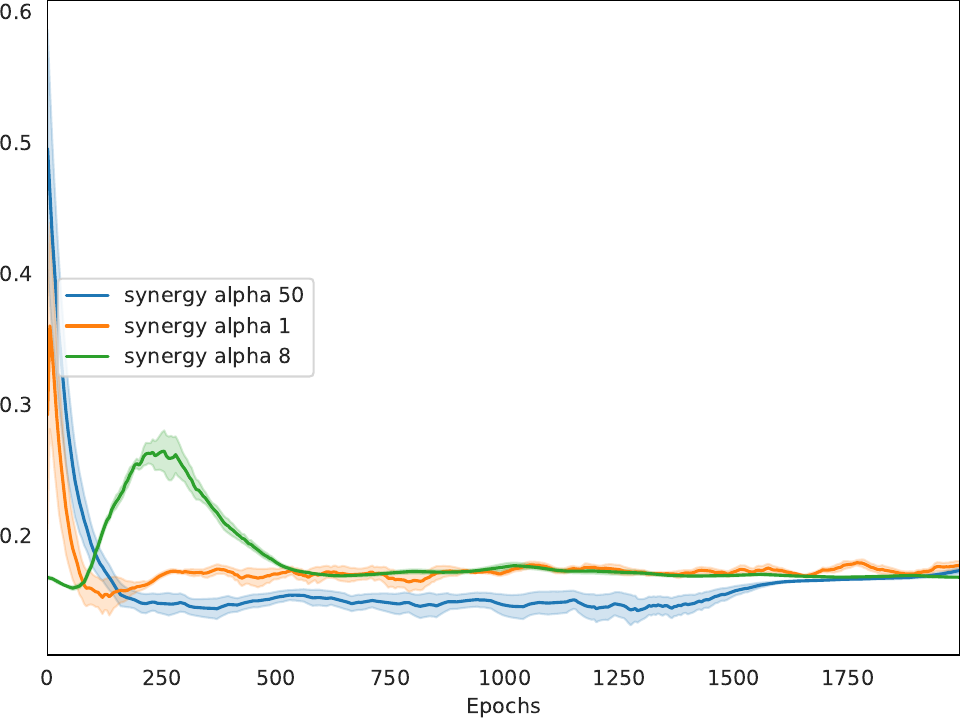}}
\caption{Left: synergy for weight decay (0.1, 2, 10, 50) Right: synergy for alpha (1, 8, 50)}
\label{peak}
\end{center}
\vskip -0.2in
\end{figure}

\section{Are the emergent synergistic sub-networks causally related to generalization?}
This experiment investigates whether sub-networks with high synergy during the emergence phase contribute to delayed generalization in models. We focus on models trained with high weight decay (2.0) and weight initialization (8). 

First, we identify and extract sub-networks with high synergy during the emergence phase. We then train these sub-networks in isolation, setting all other neurons to zero. We compare the test accuracy of these isolated sub-networks with both the original test accuracy and the accuracy of their inverse sub-networks. For comparison, we also evaluate sub-networks identified during the delayed emergence phase of a model trained with low weight decay (0.1).

\begin{figure}[ht]
\begin{center}
\centerline{\includegraphics[scale=0.25]{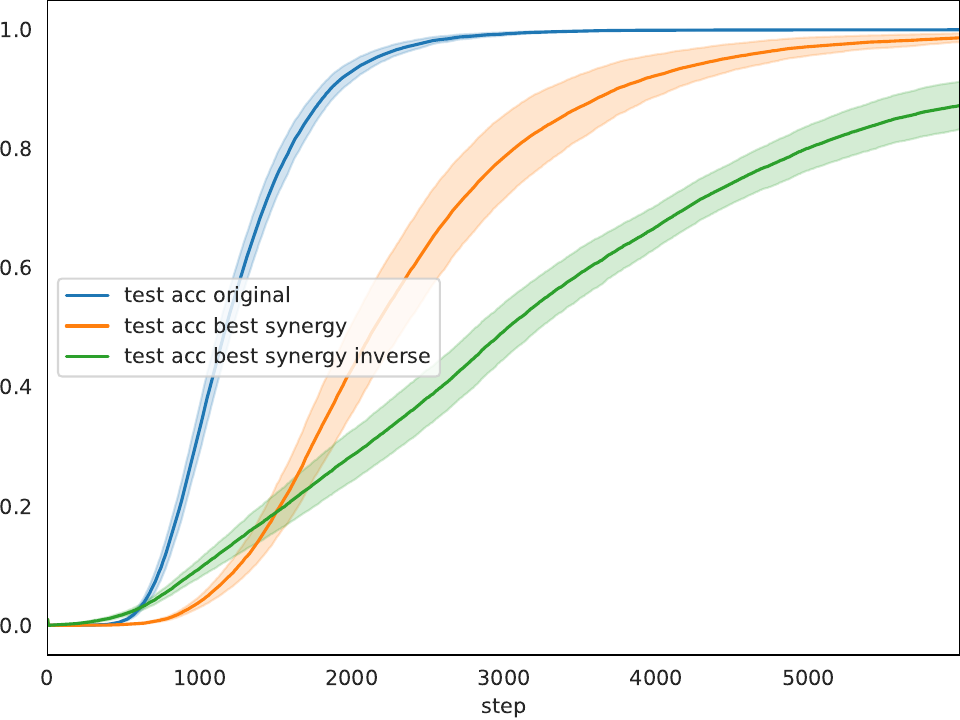}
            \includegraphics[scale=0.25]{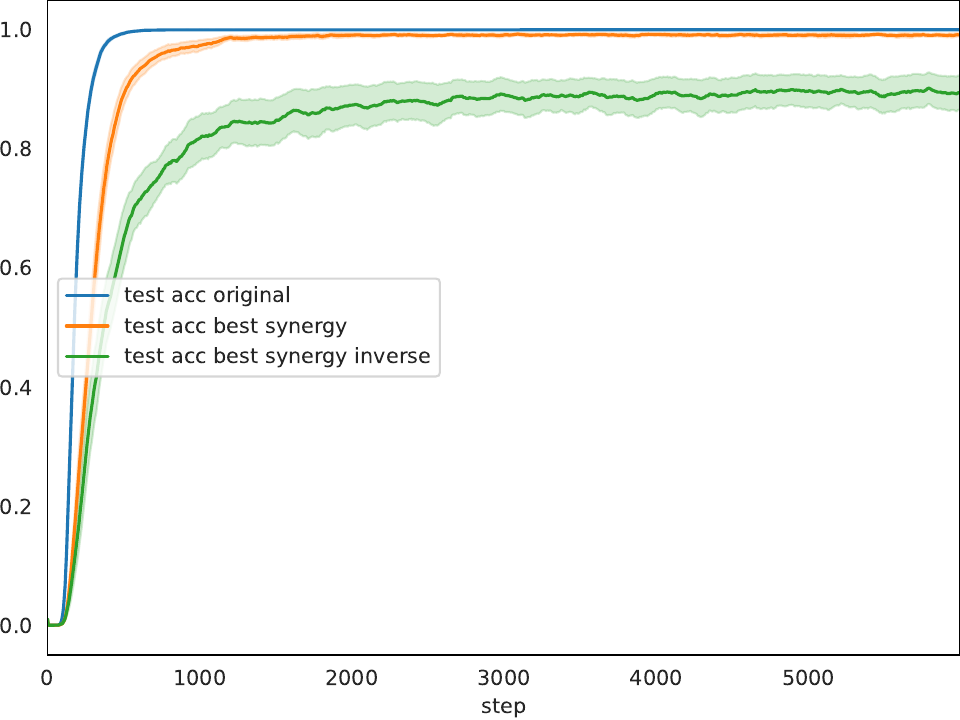}}
\caption{Left: test accuracy for synergistic sub-network with weight decay 0.1, Right:  test accuracy for synergistic sub-network with weight decay 2.0}
\label{lth_wd}
\end{center}
\vskip -0.4in
\end{figure}

\textbf{Role of weight decay} From figure \ref{lth_wd} left, we find that the synergistic network for low weight decay 0.1 achieves similar performance to the original model but the peak in test accuracy is less sudden indicating the model relies less on emergent interactions. On the other hand, the inverse model 

In contrast, figure \ref{lth_wd} right shows that synergistic sub-network for weight decay 2.0 groks in fewer epochs. We hypothesize this is due to an emergent phase as indicated by a sharper test accuracy peak and higher synergy. 

\begin{figure}[ht]
\begin{center}
\includegraphics[scale=0.25]{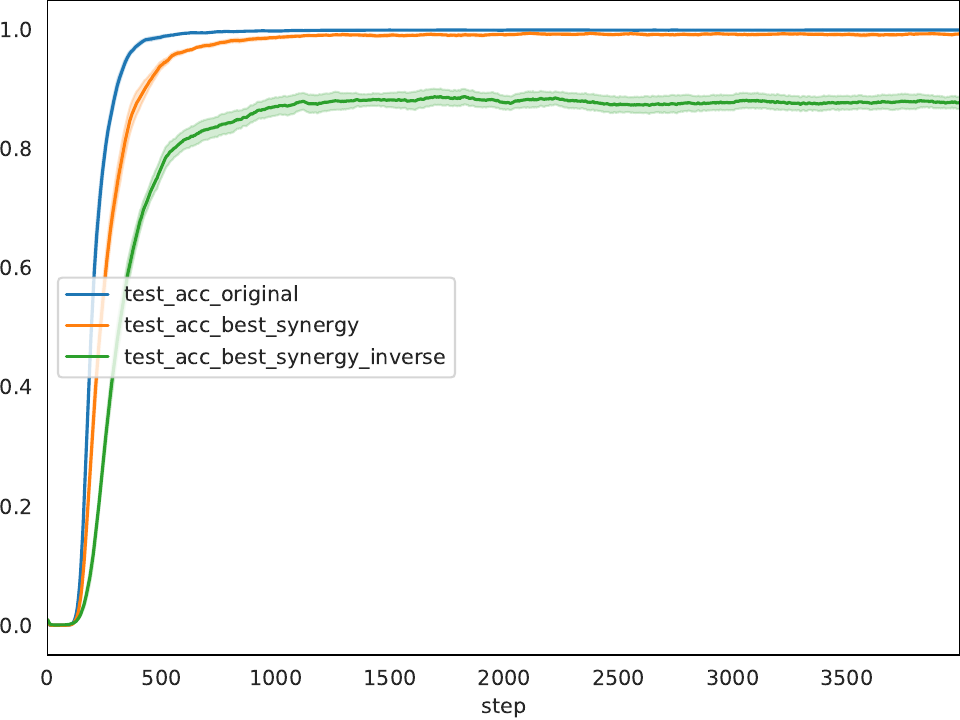}
\caption{Left: test accuracy for synergistic sub-network with weight decay 0.1, Right:  test accuracy for synergistic sub-network with weight decay 2.0}
\label{lth_alpha}
\end{center}
\end{figure}

\textbf{Role of initialization} Figure \ref{lth_alpha} shows that weight initialization 8 results in a similar grokking period than the original model. It should be noted that we observe similar results for the models with an emergent phase. These findings strengthen our belief in synergistic interactions and suggest that grokking may result from improper hyperparameter tuning.

These preliminary findings indicate that the synergistic sub-networks are causally related to generalization. However, for the models with an emergent phase, the contrast between the synergistic sub-networks and its inverse is not that large. We argue that alternative methods are needed to provide a more accurate estimate of the clusters to identify the sub-networks.

\section{Limitations}
A key limitation of our work is the simplicity of the model and benchmark. To validate the generality of the emergent phase transitions, we aim to extend this work to 1) realistic architectures based on transformers \cite{nanda_progress_2023}, or toy encoder-decoder models \cite{doshi_grok_2023} if scalability of our method is an issue. This allows us to disentangle the memorizing from the generalizing solution to better understand if these models learn isolated sub-networks, 2) evaluate alternative toy benchmarks, including sparse parity \cite{barak_hidden_2023}, XOR \cite{xu_benign_2023}, regression \cite{kumar_grokking_2023}, realistic benchmarks like MNIST to study the impact of representation learning \cite{liu_omnigrok_2022} and settings that can control grokking to increase the variety of models \cite{miller_grokking_2023}, 3) scaling the width and depth of our models. We hypothesize this might provide us insights if the relationship between grokking and double descent can be explained via synergy and redundancy \cite{huang_unified_2024, nakkiran_optimal_2021}, and 4) explore the role of optimization using gradient descent as our results might be influenced due to the internal regularization of AdamW \cite{loshchilov_fixing_2018}.


Another limitation is the difficulty of scaling the O-Information to larger networks. Our method relies on agglomerative clustering to group similar features into bins which may not fully capture the structure between features. We aim to explore alternative similarity measures such as spectral clustering \cite{hod_quantifying_2022} and the Hessian of the loss \cite{lange_clustering_2022}. 

\section{Conclusion and Discussion}
In this work, we present - to our knowledge - the first method to quantify emergent properties in neural networks for explaining grokking without relying on supervision, heuristics, or task-specific quantities. Our findings reveal an emergent phase transition where the model collectively combines features to solve arithmetic tasks. We illustrate that synergy peaks early during training could predict grokking. 

Through this preliminary work, we advocate for a paradigm shift in interpretability research from decomposing networks into the sum of isolated components (reductionism) to recognizing the emergent properties that arise from interactions between these components (emergentism). We emphasize the importance of developing methods that can quantify these higher-order interactions, as they are crucial for understanding training dynamics, generalization, and interpretability. As this is ongoing work, we conclude with future research directions. 

\textbf{Loss Geometry} Our experiments illustrated that grokking might be caused by the difficulty of optimization to navigate the loss landscapes. We plan on further exploring this by visualizing these landscapes for each phase \cite{li_visualizing_2018} and exploring the connectivity between optima \cite{garipov_loss_2018}. This might give insights on geometric properties like sharp minima giving rise to generalization \cite{hochreiter_flat_1997, keskar_large-batch_2017}. 

\textbf{Synergistic Dropout} Is the synergy between neurons a necessary and/or sufficient condition for grokking? To study if synergy causes grokking we aim to regularize for the synergy via dropout of the synergistic subnetworks \cite{srivastava_dropout_2014}. 


\textbf{Phase Transitions} Recent work observed distinct phases during optimization when training neural networks \cite{kalra_phase_2023} related to the edge of stability \cite{cohen_gradient_2020}. We aim to investigate if these phases can be related via the training dynamics to the emergent phases in this work. This might explain why previous work on grokking discovered oscillation in the loss \cite{notsawo_predicting_2023}.

\section{Author Contributions}
Kenzo carried out the experiments and wrote the paper. Daniele and Sebastiano supervised the project. 
\bibliography{references}
\bibliographystyle{icml2024}

\newpage
\appendix
\onecolumn
\section{Training results for baseline model with weight decay}

\begin{figure}[ht]
\vskip 0.1in
\begin{center}
\centerline{\includegraphics[scale=0.4]{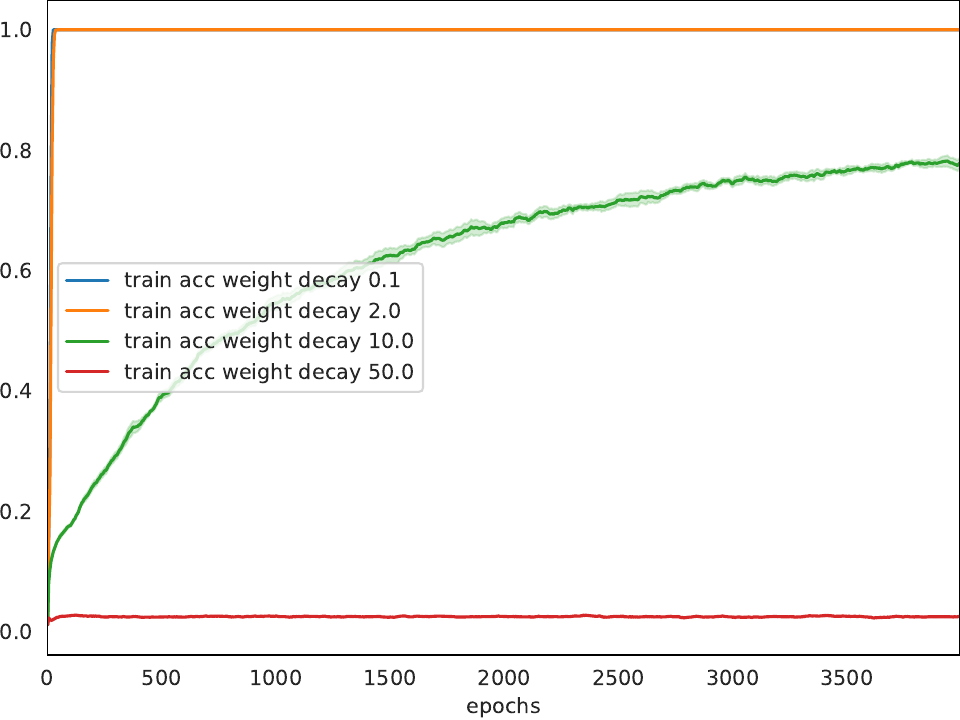}
            \includegraphics[scale=0.4]{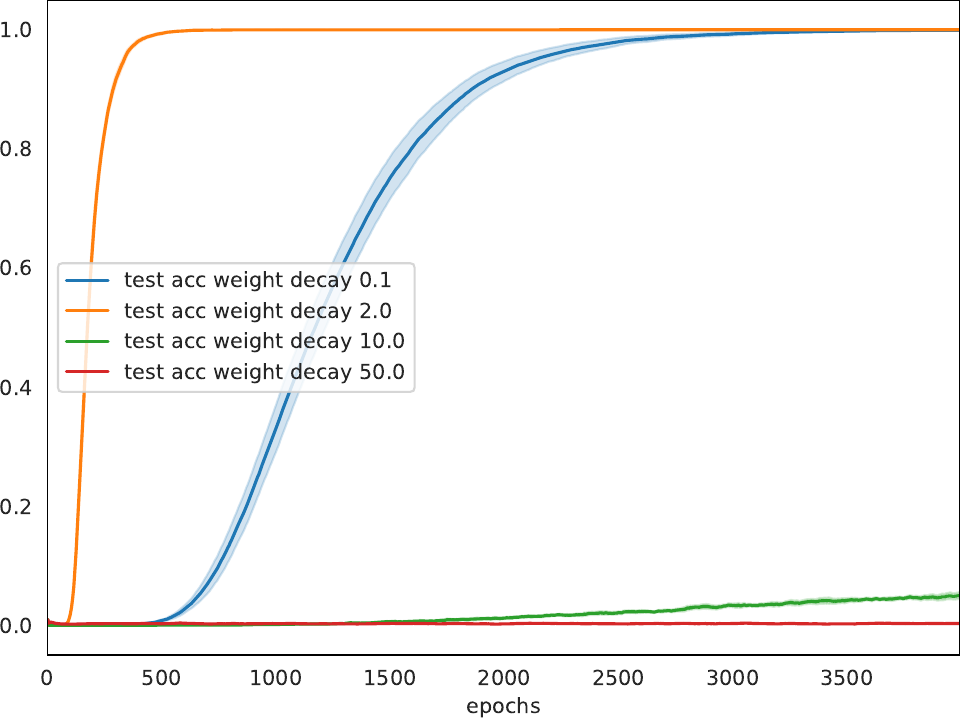}}
\caption{Left: train accuracy for weight decay (0.1, 2, 10, 50) Right: test accuracy for weight decay (0.1, 2, 10, 50)}
\label{extra_weight_decay}
\end{center}
\vskip -0.2in
\end{figure}

\section{Training results for baseline model with weight initialization alpha}

\begin{figure}[ht]
\vskip 0.1in
\begin{center}
\centerline{\includegraphics[scale=0.4]{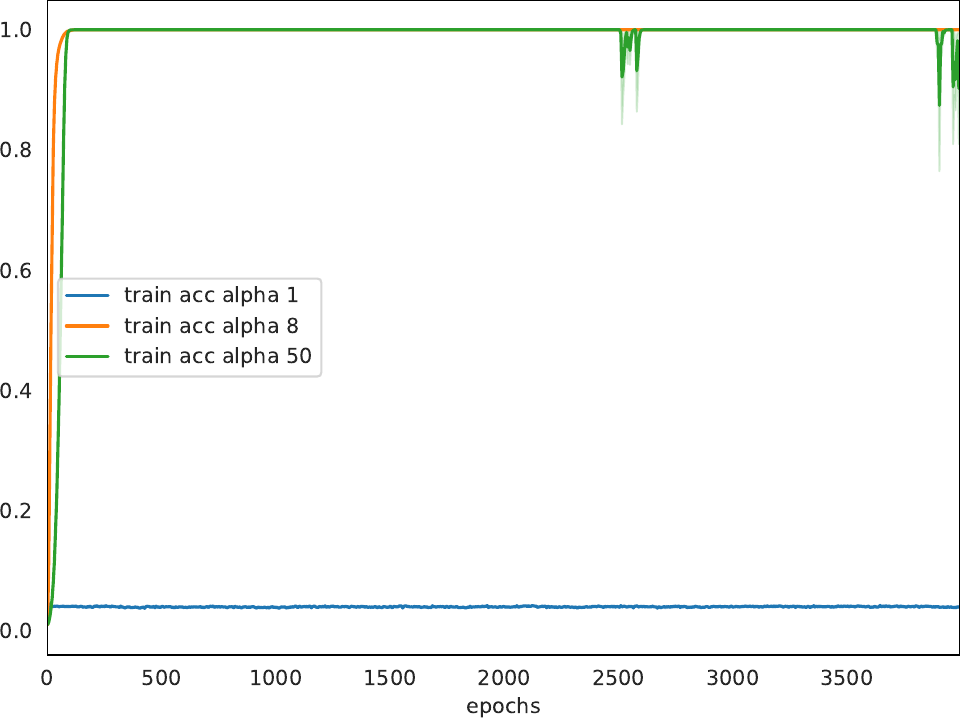}
            \includegraphics[scale=0.4]{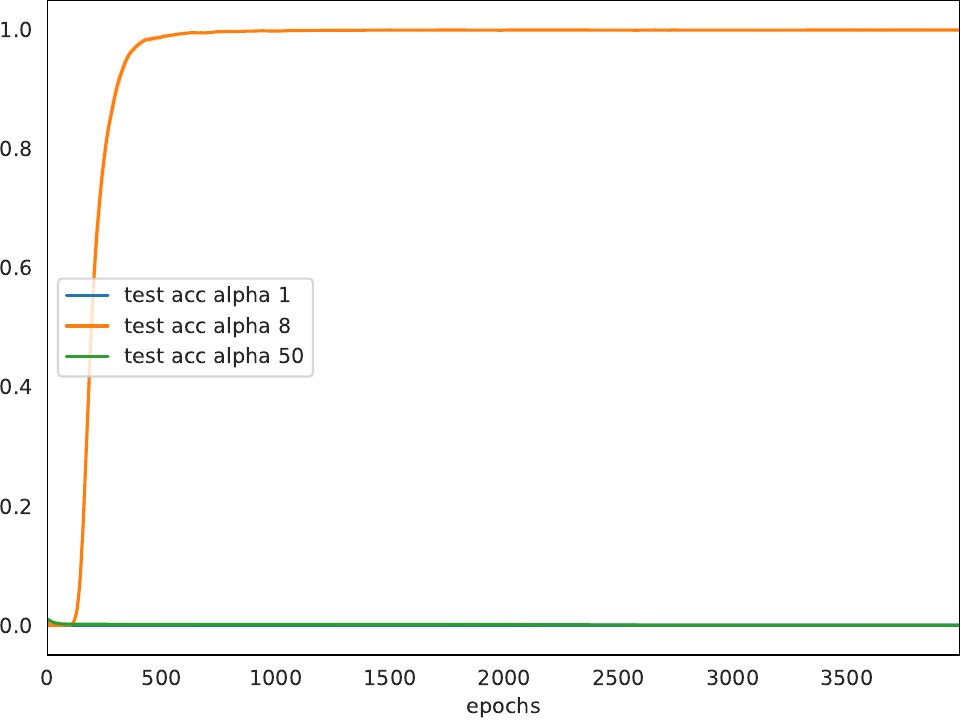}}
\caption{Left: train accuracy for alpha (1, 8, 50) Right: test accuracy for alpha (1, 8, 50)}
\label{extra_weight_init}
\end{center}
\vskip -0.2in
\end{figure}

\end{document}